\documentclass[aps,prl,twocolumn,superscriptaddress]{revtex4}
\usepackage{amsfonts,amssymb,amsmath,latexsym,epsfig}
\usepackage[usenames]{color} 
\usepackage[sort&compress]{natbib}
\begin{document}

\title{Beyond word frequency:\\ Bursts, lulls, and scaling in the temporal distributions
  of words
}

\author{Eduardo G. Altmann}
\affiliation{Northwestern Institute on Complex Systems, Northwestern University, Evanston,
 IL 60208, USA} 

\author{Janet B. Pierrehumbert}
\affiliation{Northwestern Institute on Complex Systems, Northwestern University, Evanston,
 IL 60208, USA}
\affiliation{Department of Linguistics, Northwestern University, Evanston, IL 60208, USA}

\author{Adilson E. Motter}
\affiliation{Northwestern Institute on Complex Systems, Northwestern University, Evanston,
 IL 60208, USA} 
\affiliation{Department of Physics and Astronomy, Northwestern University, Evanston, IL 60208, USA}

\begin{abstract}
{\bf Background:} Zipf's discovery that word frequency distributions obey a power law established parallels
between biological and physical processes, and language, laying the groundwork for a
complex systems perspective on human communication. 
More recent research has also identified scaling regularities in the dynamics underlying the successive occurrences
of events, suggesting the possibility of similar findings for language as well.

{\bf Methodology/Principal Findings:} By considering frequent words in USENET discussion
groups and in disparate databases where the language has different levels of formality,
here we show that the distributions of distances between successive occurrences of the
same word display bursty deviations from a Poisson process  and are well characterized
by a stretched exponential (Weibull) scaling. The extent of this deviation depends
strongly on semantic type -- a measure of the logicality of each word -- and less strongly on frequency. We develop a generative model of this behavior that fully determines the dynamics of word usage.

{\bf Conclusions/Significance:} Recurrence patterns of words are well described by a stretched exponential distribution of recurrence times, an empirical scaling that cannot be anticipated from Zipf's law.  Because the use of words provides a uniquely  precise and powerful lens on human thought and activity, our findings also have implications for other overt manifestations of collective human dynamics. 

{\bf Cite as: PLoS ONE 4 (11): e7678 (2009), doi:10.1371/journal.pone.0007678}

\end{abstract}

\maketitle

\section{Introduction}

Research on the distribution of time intervals between successive
occurrences of events has revealed correspondences between natural
phenomena on the one hand~\cite{earthquakes,bunde} and social activities on the other
hand~\cite{barabasi,politi,malmgren}.   
These studies consistently report bursty deviations both from random and from regular temporal 
distributions of events~\cite{barabasi.epl}. Taken together, they suggest the existence of a
dynamic counterpart to the universal scaling laws in magnitude and frequency
distributions~\cite{zipf,zipf2,simon,baayen,newman}.      
Language, understood as an embodied system of representation and
communication~\cite{goodwin}, is a particularly interesting and promising domain for
further exploration, because it both epitomizes social activity, and provides a medium for
conceptualizing natural and biological reality.

\begin{figure*}[!ht] 
\includegraphics[width=1.95\columnwidth]{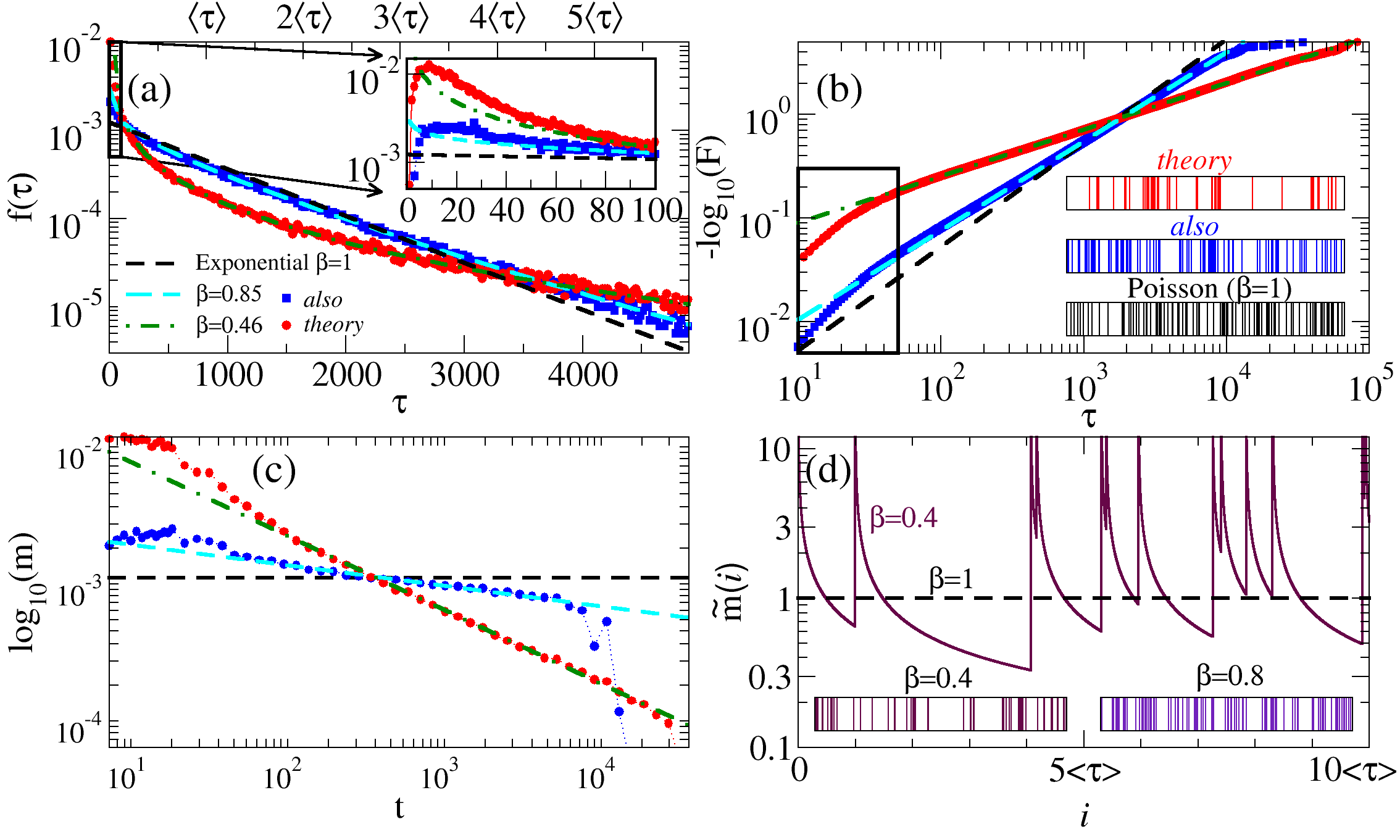}
\caption{ Recurrence time distributions for the words {\it theory} (red) and {\it
   also} (blue) in the USENET group talk.origins, a discussion group about evolution and creationism.
    Both words have a mean recurrence time of
 $\langle \tau \rangle \approx 820$. (a) Linear-logarithmic representation of $f(\tau)$, showing
 that the decay is slower than the exponential $\beta=1$ prediction~(\ref{eq.poisson}) (black
 dashed line) and follows closely
 the stretched exponential distribution~(\ref{eq.stretched}) with $\beta=0.46$
 ($R^2=0.9984$) for {\it theory} and  $\beta=0.85$ ($R^2=0.9999$)
 for {\it also}. For comparison, $\beta=1$ yields $R^2=0.49$ for the
   word {\it theory} and $R^2=0.9904$ for the word {\it also} 
   (see Text S1, {\em Fitting Procedures}).
The inset in (a) shows a magnification for short times. A word-dependent peak at~$\tau<50$
reflects the domination of syntactic effects and local discourse structure at this scale.
 (b) Cumulative distribution function~$F(\tau)$ in a scale in which the stretched exponential~(\ref{eq.stretched})  appears as
 a straight line. The
   panels in the inset show $100$ occurrences (top to bottom): of the word {\it theory}, of the word {\it
   also}, and of a randomly distributed word ($\beta=1$).
(c) The probability
of word usage $m(t)$ for the words {\it theory} and {\it also}. The data are binned
logarithmically and the straight lines correspond to Eq.~(\ref{eq.beta}).
(d) Illustration of the generative model 
 for the usage of individual words when~$\beta=0.4$, where the spikes indicate the times
 at which the word is used. The probability~$\tilde{m}(i)$ of using a word
 decays as a piece-wise power-law function since its last use, as determined by
 Eq.~(\ref{eq.beta}). The Poisson case corresponds to constant~$\tilde{m}$. The panels at the bottom show $100$ occurrences of words
 generated by the model for~$\beta=0.4$ and~$\beta=0.8$. 
}
\label{fig1}
\end{figure*}

The fields of statistical natural language processing and psycholinguistics study
language from a dynamical point of view. Both treat language processing as
encoding and decoding of information. 
In psycholinguistics, 
the local likelihood (or predictability) of words is a central focus of current
research~\cite{bell}. Many widely used practical applications of statistical 
natural language processing, such as document retrieval based on keywords, also exploit
dynamic patterns in word statistics~\cite{baayen,church,katz}.
Particularly important for these applications, and also noticed in different
contexts~\cite{montemurro,ortuno,herrera,sarkar,eckman,serrano},
is the non-uniform distribution of content words through a text,
suggesting that 
connections to the previous discoveries about inter-event distributions may be revealed 
through a systematic investigation of the recurrence times of different words.

With the rise of the Internet, large records of spontaneous and collective language are
now available for scientific inquiry~\cite{watts,wu_hub,lambiotte}, 
allowing statistical questions about
language to be investigated with an unprecedented precision. At the same time,
large-scale text mining and document classification is of ever-increasing importance~\cite{nigam}.
The primary datasets used in our study are USENET discussion groups available through Google
(http://groups.google.com). These exemplify spontaneous linguistic interactions in large
communities over a long period of time.
We first focus on the $N = 2,128$ words that  occurred more than $10,000$ times  between
Sept.\ 1986 and Mar.\ 2008 in a ($2\;10^8$-word) discussion group, talk.origins. The
data were collated chronologically, maintaining the thread structure 
(see Text S1, {\em Databases}).

Here, we show that long-time word recurrence patterns follow a stretched exponential
distribution, owing to bursts and lulls in word usage.
We focus on time scales that exceed the scale of {\em syntactic} relations, and the burstiness of the words
is driven by their semantics (that is, by what they mean). The burstiness of physical events and socially
contextualized choices makes words more bursty than an exponential distribution. However, we
show that words are 
typically less bursty than 
other human activities~\cite{vazquez} due to their {\em logicality} or  {\em
permutability}~\cite{benthem, vonFintel}, technical constructs of formal semantics that
index the extent to which the meanings and usage of words are stable over changes in the discourse
context. Our quantitative analysis of the empirical data confirms the inverse relationship
between burstiness and permutability. The model we develop to explain these observations shares the generative spirit of local ($n$-gram)
and weakly non-local models of text classification and
generation~\cite{shannon,grosz,ron}. However it focuses on long time-scales, picking
up at temporal scales where studies of local predictability and coherence leave
off~\cite{bell}. 
We verify the generality of our main findings using different databases, 
including books of different genres and a series of political debates.

\section{ Results and Discussion}

We are interested in the temporal distribution of each word~$w$. 
All words   are enumerated in order of appearance, $i = 1, 2, ..., N$, 
where $i$ plays the role of the time along the text. 
The recurrence time $\tau^w_{j}=i^w_{j+1}-i^w_{j}$
is defined by the number of words between two successive 
uses ($i^w_j$ and $i^w_{j+1}$) of word~$w$ (plus one). For instance, the first appearances of the word {\it the} in the abstract above
 are at  $i^{the}_1=22, i_2^{the}=41, i_3^{the}=44,
i_4^{the}=50,...$, leading to a sequence of recurrence times $\tau^{the}_1= 19, \tau^{the}_2= 3, \tau^{the}_3= 6,...$.
We are interested in the distribution~$f_w(\tau)$ of $\tau=\tau^w_{j}$, $j=1,...,N_w$. 
The mean recurrence time, called 
by Zipf the wavelength of the word~\cite{zipf}, is
given by~$\langle \tau^w \rangle = N/ N_w \equiv 1/\nu_w$~\cite{bunde}  (hereafter we drop~$w$ from our
notation). 
It is mathematically convenient to consider $\tau$ to be a continuous time variable (an
assumption that is
justified by our interested in $\tau \gg 1$) and to use the
cumulative probability density function defined by $F(\tau) \equiv 
\int_\tau^\infty f(\tilde{\tau})d\tilde{\tau}$, which satisfies $F(0)=1$ and $ 
\int_0^\infty F(\tau) d\tau = \int_0^\infty \tau f(\tau) d\tau = \langle \tau \rangle = 1/\nu$.

The first point of interest is how the distribution~$f(\tau)$ [or $F(\tau)$] deviates
from the exponential distribution 
\begin{equation}\label{eq.poisson}
f_P(\tau)=\mu e^{-\mu \tau}, \;\;\;\; F_P(\tau) = e^{-\mu \tau},
\end{equation}
where $\langle \tau \rangle = 1/\nu$ leads to $\mu=\nu$. 
The exponential distribution is predicted by a simple {\it bag-of-words} model in which the
probability~$\mu$ of using the word is time independent and
equals~$\nu$ (a Poisson process with rate~$\mu=\nu$)~\cite{shannon,nigam,church,katz,sarkar}, as observed if the words 
in the text are randomly permuted.
Deviations are caused by the way that people choose their words in context. Numerous
studies, as reviewed in Ref.~\cite{tanenhaus}, already demonstrate that the language users
dynamically modify their 
use of nouns and noun phrases as a function of the linguistic and external context.
We analyze such modifications for all types of words. 

Figure~\ref{fig1} shows the empirical results obtained for the example words {\it theory}
and {\it also} 
in the talk.origins group of the USENET database.
Both words have $\langle \tau \rangle \approx 820$ but are linguistically 
quite different: while {\it theory} is a common noun, {\it also} is an 
 adverb that functions semantically as an operator.
The deviation from the Poisson prediction~(\ref{eq.poisson}) is apparent in Fig.~\ref{fig1}(a-c):
$f(\tau)$ is larger than the exponential distribution for distances~$\tau$ both much shorter 
and much longer than~$\langle \tau \rangle$, while it is smaller for~$\tau\approx \langle \tau
\rangle$. 
Both words exhibit a most probable recurrence time $\tau\lesssim 20$ 
and a monotonically decaying distribution $f(\tau)$ for larger times [Fig.~\ref{fig1}(a)].
Comparing the 
insets in Fig.~\ref{fig1}(b), one sees that the occurrences
of {\it theory} are clustered close to each other in a phenomenon known as
burstiness~\cite{barabasi.epl,church,katz,sarkar,serrano}.
Due to burstiness, the  frequency of the word
{\it theory}  estimated from a small sample would differ a great deal as a function of
exactly where the sample was drawn.
Similar but lesser deviations are observed for the word
{\it also}.

\begin{figure*}[ht!] 
\includegraphics[width=1.95\columnwidth]{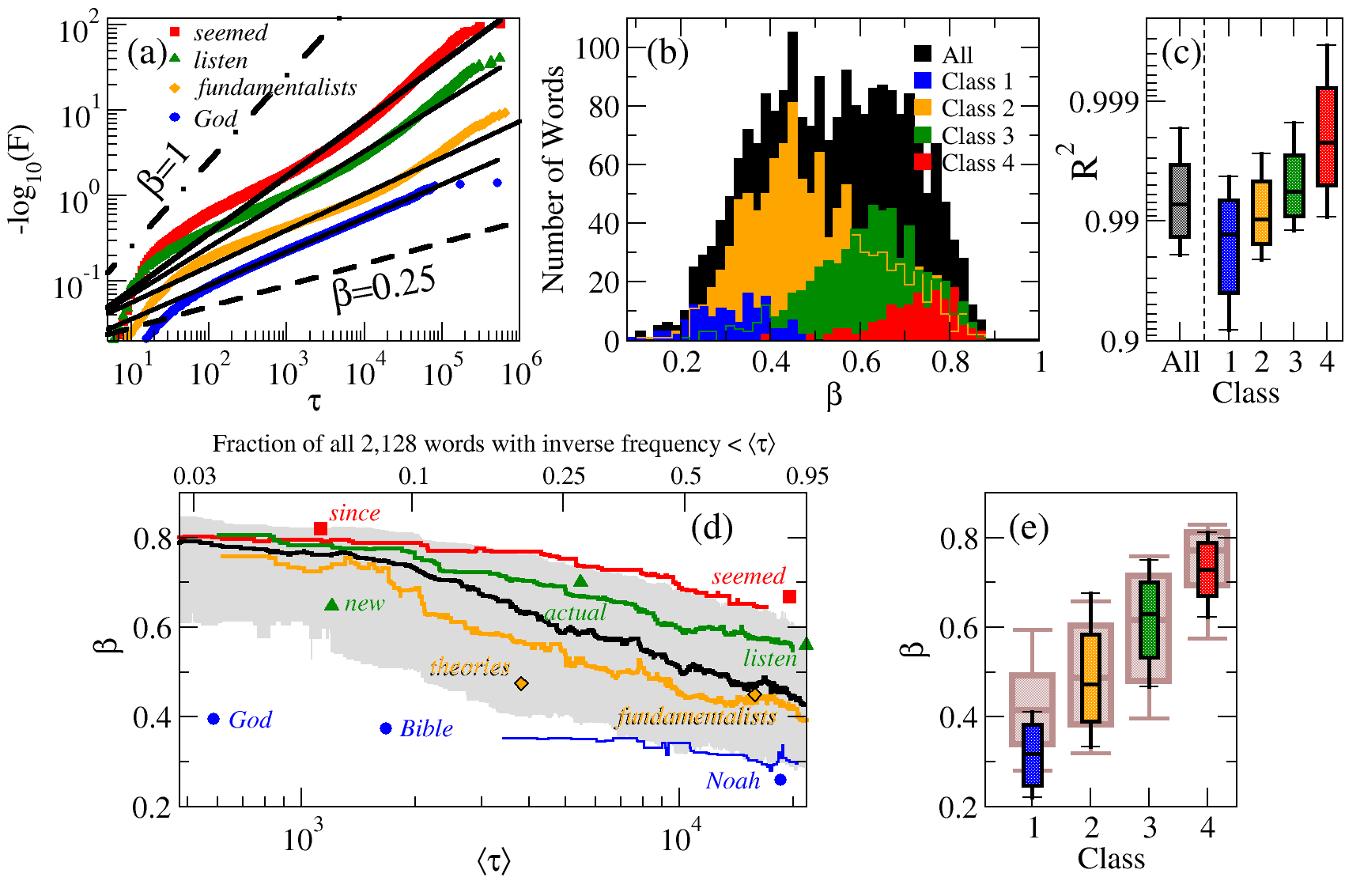}
\caption{ Dependence of $\beta$ on semantic Class and frequency for the $2,128$ most frequent
 words of the USENET group talk.origins.  Different classes of words (see Table~\ref{tab.classes}) 
 are marked in different colors. 
(a) Fitting  
of~$\beta$ exemplified for four words with $R^2\approx
 R^2_{median}=0.993$ (bottom to top):  {\it God}, Class 1,
 $\beta=0.39, \langle \tau \rangle = 586$; {\it fundamentalists}, Class 2,
 $\beta=0.45, \langle \tau \rangle = 15,825$;  {\it listen}, Class 3, $\beta=0.56,
 \langle \tau \rangle = 21,971$; {\it seemed}, Class 4, $\beta=0.67, \langle \tau \rangle =
 19,564$.
 (b)
 Histogram of the   fitted~$\beta$, providing evidence that the Class is determinant 
 to  the value of~$\beta$.
(c) Quality of fit quantified in terms of the coefficient of  determination~$R^2$
 between the fitted stretched exponential and the empirical~$F(\tau)$ (see Text S1, {\em  Quality of
 Fit}). 
The box-plots are centered at the median and indicate the $1,2,6,7$ octiles. 
For comparison, an exponential fit with two free parameters yields~$R^2_{median}=0.907$
(see Text S1, {\em Deviation from the Exponential Distribution}). 
(d)  Relative dependence of $\beta$ on Class and $\langle \tau \rangle = 1/\nu$ (inverse
frequency), indicating: running median on words ordered according to~$\langle \tau \rangle$ (center
black line) and $1$-st and~$7$-th octiles (boundaries of the gray region); and running medians on
words by Class (colored lines, Class 1-4, from bottom to top) with illustrative words for
each Class. At each~$\langle \tau
\rangle$, large variability in~$\beta$ and a systematic ordering by Class is observed.
(e) Box-plots of the variation of $\beta$ for words in a given Class. 
The box-plots
 in the background are obtained using frequency to divide all words in four groups with
 the same number of words of the semantic Classes (first box-plot has words with lowest frequency and last box-plot has words
 with highest frequency). The classification based on Classes leads to a narrower
 distribution of~$\beta$'s inside Class and to a better discrimination between Classes.
}
\label{fig2}
\end{figure*}

Central to our discussion,
Fig.~\ref{fig1} shows that the distributions of both 
words can be well described by the single free parameter~$\beta$
of the stretched exponential distribution 
\begin{equation}\label{eq.stretched}
f_\beta(\tau) = a \beta \tau^{\beta-1}e^{-a \tau^\beta}, \;\;\;\; F_\beta(\tau) = e^{-a \tau^\beta} ,
\end{equation}
where $a = a_\beta =(\nu \; \Gamma(\frac{\beta+1}{\beta})]^\beta$ is obtained by
imposing~$\langle \tau \rangle = 1/\nu$, $\Gamma$ is the Gamma function, 
and~$0< \beta \leq 1$. 
Distribution~(\ref{eq.stretched}), also known as Weibull distribution, 
and similar stretched exponential distributions describe a variety of
phenomena~\cite{barabasi.epl,wu_hub,sornette,cox,redner}, including the 
recurrence time between extreme events in time series with long-term
correlations~\cite{bunde,kantz}. 
The stretched exponential~(\ref{eq.stretched}) 
is more skewed than the simple exponential 
distribution~(\ref{eq.poisson}), 
which corresponds to the limiting case~$\beta=1$, but less skewed than a
power law, 
which is approached for~$\beta\rightarrow0$. 

A crucial test
for the claim that an empirical distribution~$F(\tau)$ follows a stretched
exponential~$F_\beta$ is to represent $-\log(F(\tau))$ as a function of $\tau$ in a
double logarithmic plot~\cite{bunde}.  
The straight line behavior for almost three decades
shown in Fig.~\ref{fig1}(b), 
which is
illustrative of the words in our  
datasets, provides strong evidence for the stretched exponential scaling
(spam-related deviations for long~$\tau$ are discussed in Text S1, {\em Databases}). 
This is a clear advance over the closest precedents to our results: 
(i) In Ref.~\cite{zipf2} Zipf proposed a power-law decay, which would appear as an horizontal line in
Fig.~\ref{fig1}b. 
(ii) Refs.~\cite{church,katz} compare two non-stationary Poisson processes for predicting the counts of words
in documents (see Text S1, {\em Counting Distribution}); 
(iii) Ref.~\cite{sarkar} proposes a non-homogeneous Poisson process for recurrence times,
using a mixture of two exponentials with a total of four free parameters;
(iv) Ref.~\cite{hrebicek} uses the Zipf-Alekseev distribution~$f(\tau) \sim \tau^{-\alpha-b\ln(\tau)}$,
which we found to underestimate the decay rate for large~$\tau$ and to leave larger residuals than our
fittings (see Text S1, {\em Zipf-Alekseev Distribution}). 
The stretched exponential distribution was found to describe the time between usages of words in Blogs
 and RSS feeds in Ref.~\cite{lambiotte}. However, time was measured as actual time and
 the same distribution was found for different types of words, suggesting that
their observations are driven by the bursty update of webpages, a related but different effect. 
More strongly related to our study is Ref. \cite{malmgren}'s analysis of email activity, in which a
non-homogeneous Poisson process captures the way one email can trigger the next.\\

\noindent {\bf Generative Model.} Motivated by the successful description of the stretched exponential
distribution~(\ref{eq.stretched}), we search for a generative stochastic 
process that can model word usage.
We consider the inverse frequency~$\langle \tau \rangle$ as given
and focus on describing
how the words are distributed throughout the text.
We assume that our text (abstractly regarded as arbitrarily long) 
is generated by a
well-defined stationary stochastic process with finite~$\langle \tau \rangle$ for 
the words of interest.  We further assume that the probability~$m(t)$ of using
the word~$w$ depends only on the distance~$t$
since the last occurrence of the word. The latter means that we are modeling the word 
usage as a {\em renewal process}~\cite{cox,kantz}.
The distribution of recurrence times is then given by the (joint)
probability of having the word at distance~$\tau$ and not having this word 
for $t<\tau$:
$$f(\tau) = m(\tau) \prod_{i=1}^{\tau-1} (1-m(i)) \approx m(\tau) e^{-\int_0^\tau m(t)dt}.$$
The cumulative distribution function is written as
\begin{equation}\label{eq.rhomut}
F(\tau) = e^{-\int_0^\tau m(t) dt}.
\end{equation}
The time dependent probability~$m(t)$, 
also known as {\em hazard function}, can
 be obtained empirically as~$m(t)=f(t)/F(t)$  (see Text S1, {\it Hazard Function}). 
Equation~(\ref{eq.rhomut}) reduces to the exponential
distribution~(\ref{eq.poisson}) for a time independent probability $m(t) = \mu = 1/\langle
\tau \rangle$.
The stretched exponential distribution~(\ref{eq.stretched}) is obtained
from~(\ref{eq.rhomut}) by asserting that~\cite{cox,kantz,mcshane}
\begin{equation}\label{eq.beta}
m(t) = a \beta t^{-(1-\beta)} \;\; \mbox{for}\;\;  0<\beta\leq 1. 
\end{equation}
This assertion means that in our model, the probability of using a word decays as a power
law since the last use of that word. 
This is further justified by the power-law behavior of~$m(t)$ determined directly from the
empirical data, as shown in 
Fig.~\ref{fig1}(c) and Text S1, Fig.~9, 
and is in agreement with results from
mathematical psychology~\cite{wixted,anderson} and information retrieval~\cite{anderson}.
The Weibull renewal process we propose can be analyzed formally as a particular instance of a doubly stochastic Poisson process~\cite{yannaros}.

Our model is illustrated in Fig.~\ref{fig1}(d)
and can be interpreted as 
a bag-of-words with memory that accounts for the burstiness of word usage. 
This model does not reproduce the  positive  correlations between 
$\tau_j$ and $\tau_{j+p}$~\cite{bunde,barabasi.epl,eckman}, which are usually small (less than $20$\% for~$p=1$) but 
decay slowly with~$p$ (see Text S1, {\em Correlation in $\{\tau_j\}$} ). 
These correlations quantify the extent to which the renewal model is a
 good approximation of the actual generative process, and show that the burstiness of words exists
 not only as a departure of~$f(\tau)$ from  the exponential distribution, but also as a clustering
 of small (large) $\tau$~\cite{barabasi.epl} (see Text S1, {\em Independence of
   $\{\tau_j\}$}). 
The advantage of the renewal description is that the model (i)  can be substantiated to a vast 
literature describing power-law decay of memory in agreement with Eq.~(\ref{eq.beta}), see
Refs.~\cite{wixted,anderson} and references therein, and (ii) fully determines the
dynamics (allowing, e.g., the precise derivation of counting distributions~\cite{mcshane},
which are
used in applications to document classification~\cite{church,katz} and information
retrieval~\cite{anderson}). \\

\noindent {\bf Word Dependence.} We have seen in Fig.~\ref{fig1} that the word-dependent deviation
from the exponential distribution is encapsulated in the parameter~$\beta$: the smaller
the~$\beta$ for any given word, the larger the deviation (see Text S1,
{\em Deviation from the  Exponential Distribution}). 
Next we investigate the
dominant effects that determine the value of the parameter~$\beta$ of a word.  
Previous research has observed that frequent {\em function} words 
 (such as conjunctions and determiners)
usually are closer to the random (Poisson) prediction while less frequent {\em content}
words (particularly names and common nouns)  are more bursty. These observations were quantified using: (i) an entropic analysis of
texts~\cite{montemurro}; (ii) the variance of the sequence of recurrence
times~\cite{ortuno}; (iii) the recurrence time distribution~\cite{sarkar,corral}; and (iv)
the related distribution of the number of occurrences of words per document~\cite{church,katz}.
Because we have a large database and do not bin the datastream into documents, we are able
to go beyond these insightful works and systematically examine frequency and linguistic
status as factors in word burstiness.

\begin{figure*}[ht!]
\includegraphics[width=1.95\columnwidth]{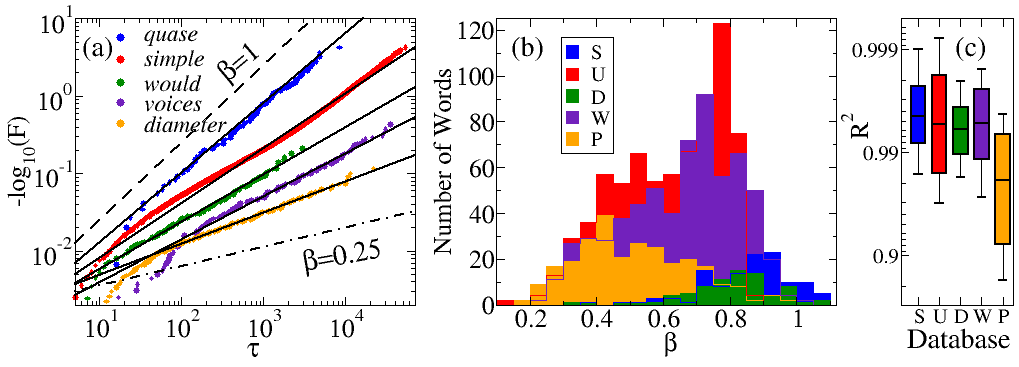}
\caption{
Stretched exponential recurrence time distributions observed in different
 databases. The databases consist of the documentary novel {\it Os Sert\~oes} by  Euclides da Cunha (S), in
 Portuguese ($N\approx 1.5\; 10^5$);
 the USENET group comp.os.linux.misc (U) between Aug. $1993$ and  
 Mar. $2008$ ($N\approx 6\; 10^7$); 
 the three Obama-McCain
 debates of the 2008 United States presidential election (D) arranged
 in chronological order ($N\approx 5\; 10^4$); 
an English edition of the novel {\it War and Peace} by Leon Tolstoy (W) ($N \approx 6\; 10^5$);
and 
the first English edition of   Isaac Newton's  {\it Principia} (P)  ($N \approx 2\; 10^5$).
All words appearing
 more than $100$ times were considered in S ($117$ words), D ($78$ words), P ($268$ words), and W
 ($633$ words), whereas in U all $733$ words appearing more than $10,000$ times were 
 used (see Text S1, {\em Databases}). 
(a) Recurrence time distributions for the words  
 {\it quase} in S ($\beta=0.88, \langle \tau \rangle = 1,204, R^2=0.996$), {\it simple} in U
 ($\beta=0.71, \langle \tau \rangle = 3,397, R^2=0.996$), 
 {\it would} in D  ($\beta=0.61, \langle \tau \rangle = 359.5, R^2= 0.995$), 
 {\it voices} in W ($\beta=0.58, \langle \tau \rangle = 3,946, R^2=0.994$), and 
 {\it diameter} in P ($\beta=0.40, \langle \tau \rangle = 1,129,  R^2=0.975$). 
  (b) Histograms of the  fitted~$\beta$ for all datasets.
Due to sample size limits, the analysis into semantic 
 Classes is not feasible for the smaller datasets. (c) Box-plots of the coefficient of
 determination~$R^2$ of the corresponding stretched exponential fit. 
}
\label{fig3}
\end{figure*}

Our large database allows a detailed analysis of words that, despite being in the same
frequency range, have very different statistical behavior.
For instance, in the range $2,000 < \langle \tau \rangle
<3,000$, words with  high $\beta$ ($\approx 0.80$) include {\it once, certainly, 
 instead, yet, give, try, makes,} and {\em seem}; the few words with $\beta \lessapprox 0.40$
include 
{\it design, selection, intelligent,} and {\em Wilkins}. Corroborating Ref.~\cite{church}, 
it is evident that words with low $\beta$ better characterize the discourse
topic. However, these examples also show that the distinction between function words and
content words cannot be explanatory. For instance, many content words, such as the adverbs
and verbs 
of mental representation
in the list just above, have $\beta$ values as high as many function words.
Here we obtain a deeper level of explanation by drawing on tools from formal
 semantics, specifically on type theory~\cite{montague,partee,benthem},
and on dynamic theories of semantics~\cite{heim,kamp}, 
which model how 
words and sentences
update the discourse context over time. 
We use semantics rather than syntax because syntax governs how words are
 combined into sentences, and we are interested in much 
 longer time scales over which syntactic relations are not defined. Type theory
 establishes a scale from simple entities (e.g., proper nouns) to high type words (e.g., 
 words that cannot be described using first-order logic, including intensional
   expressions and operators). Simplifying the technical
literature in the interests of good sample sizes and coding reliability, we define a
ladder of four semantic classes, as listed in Table~\ref{tab.classes}.

\begin{table}[h]
\begin{tabular}{c c c}
\hline
{\bf Class}
&{\bf Name}  
&{\bf Examples of words}\\
\hline
\hline
{1}
&{Entities}
&{Africa, Bible, Darwin}\\
\hline
{2}
&{Predicates and Relations}
&{blue, die, in, religion}\\ 
\hline
{3}
&{Modifiers and Operators}
&{believe, everyone, forty}\\
\hline
{4}
&{Higher Level Operators}
&{hence, let, supposedly, the}\\
\hline
\end{tabular}
\caption{ Examples of the classification of words by semantic types. The
 primitive types are entities {\it e}, exemplified by proper nouns such as {\it Darwin}
 (Class 1), and truth values, {\it t } (which are the values of sentences). Predicates or
 relations, such as 
 the simple verb {\it die}, and the adjective/noun blue, take entities as
 arguments and map them to sentences  (e.g., {\it Darwin dies}, {\it Tahoe is blue}). They
are classified as $< e, t >$ (Class 2). The notation 
 $<x,y>$ denotes a mapping  from an element $x$ in the domain to the image
 $y$~\cite{montague,partee}. The semantic types of higher Classes are established by
 assessing what mappings they perform when they are instantiated. For 
example, {\it everyone} is of type $<< e, t >, t >$ ~(Class 3), because it is a mapping from sets of
properties of entities to truth values~\cite{partee}; 
the verb {\it believe} shares this classification as a verb involving mental
 representation. The adverb
{\it supposedly} is a higher 
order operator (Class 4), because it modifies other modifiers. Following Ref.~\cite{partee}
(contra Ref.~\cite{montague}) words are coded by the lowest type in which they commonly
occur (see Text S1, {\em Coding of Semantic Types}). 
}\label{tab.classes}
\end{table}

In Fig.~\ref{fig2}, we report our systematical analysis of the recurrence time distribution of 
all $2,128$ words that appeared more than ten thousand times in our database (for word-specific results see Table S1).  
We find a wide
range of values for the burstiness parameter~$\beta$~[$0.2<\beta<0.9$, Fig.~\ref{fig2}(a,b)] and 
the stretched exponential distribution describes well most of the
words~[$R^2_{median} = 0.993$, Fig.~\ref{fig2}(c)]. 
The Class-specific results displayed in  Fig.~\ref{fig2}(a-c) show  that words of all classes are
accurately described by the same statistical model over a wide range of scales, a strong
indication of a universal process governing word usage at these scales. 
Figure~\ref{fig2}(b) also reveals a systematic dependence of $\beta$ on the semantic Classes: 
burstiness increases ($\beta$ decreases) with decreasing semantic Class. 
This relation implies that words functioning unambiguously as Class 3 verbs should be less bursty
than words of the same frequency functioning unambiguously as common nouns (Class 2). This
prediction is confirmed by a paired comparison in our database: such verbs have a higher~$\beta$
in~$103$ out of $116$ pairs of verbs and frequency-matched nouns (sign test, $P\leq 8
\; 10^{-19}$).
The relation
applies even to morphologically related forms of the same word stem (see Text S1, {\em
  Lemmatization}): 
for $37$ out of the $47$ pairs of Class 3 adjectives and Class 4 adverbs in the
database that are derived with {\it -ly}, such as {\it perfect, perfectly}, the adverbial form
has a higher $\beta$ than the adjective form (sign test, $P \leq 5 \;
10^{-5}$). 
Figure~\ref{fig2}(d) shows the dependence of~$\beta$ on inverse frequency~$\langle \tau
\rangle$. This figure may be compared to the TF-IDF (term frequency-inverse document
frequency) method used for keyword identification~\cite{church}, but it is computed from a
single document (see also Refs.~\cite{montemurro,ortuno,herrera}). Figure~\ref{fig2}(d) reveals that~$\beta$ is 
correlated with~$\langle\tau\rangle$ and that the Class ordering observed in
Fig.~\ref{fig2}(b) is valid at all~$\langle \tau \rangle$s. The detailed analysis in
Fig.~\ref{fig2}(e) demonstrates that semantic Class is more important than frequency as a
predictor of burstiness (Class 
accounts for $0.32$ and log-frequency for $0.26$ of the variance of~$\beta$, by the test proposed
in Ref.~\cite{kruskal}).

We are now in a position to discuss why burstiness depends on semantic Class. A
straw man theory would seek to derive the burstiness of  referring expressions directly
from the burstiness of their referents.  The limitations of such a theory are obvious:
{\it Oxygen} is a very bursty word in our database ($\beta \approx 0.25$) 
though oxygen is ubiquitous. A more careful observer would connect the burstiness
of words to the human decisions to perform activities related to the words. For instance,
the recurrence time between sending emails is known 
to approximately follow a power law~\cite{barabasi,malmgren}. However, in our 
database the
word {\it email} is significantly closer to the exponential
($\beta \approx 0.5)$ than a
power law would predict~($\beta \rightarrow 0$). Indeed, a defining characteristic of human language
is the ability
to refer to entities and events that are not present in the immediate
reality~\cite{hockett}. 
These nontrivial connections between language and the world are investigated in
semantics. An insight on the problem of word usage 
can be obtained from Ref.~\cite{benthem}, which establishes that the meaning and applicability of 
words with great {\em logicality} remains invariant under {\em permutations} of alternatives for
the entities and relations specified in the constructions in which they appear. Here we consider
permutability to be proportional to the semantic Classes of Table~\ref{tab.classes}.
As a long discourse unfolds exploring different constructions, we expect  
words with higher 
permutability (higher semantic Class) to be more homogeneously
distributed throughout the discourse and therefore have higher~$\beta$ (be less
bursty). 
Critical to this explanation is the fact that  human language manipulates
representations of abstract operators and mental states~\cite{hauser}.  However, the overt
statistics of recurrence times do not need to be learned word by word. It seems more
likely that they are an epiphenomenal 
result of the differential contextualization of word meanings. 
The fact that the behavior of almost all
words deviate from a Poisson  process to at least some extent,  indicates that 
the permutability and usage of almost all words are contextually restricted to some 
degree, whether by their intrinsic meaning or by their social connotations.  \\

\noindent {\bf Different Databases.} 
In Fig.~\ref{fig3} we verify our main results using 
databases of different sizes and characterized by different levels of formality. We analyzed a second example
of a USENET group (U),  a series of political debates (D), two novels (S,W), and a technical 
book (P) (for word-specific results see Table S1). The stretched exponential provides a close
fit for frequent words in these datasets [Fig.~\ref{fig3}(a,c)],~and a wide and smoothly varying range of $\beta$s is 
observed in each case [Fig.~\ref{fig3}(b)]. 
The technical book exhibits lower $\beta$ values, which
can be attributed to the predominance of
specific scientific terms.
These datasets include examples of texts differing by almost four orders of magnitudes in size, generated by a single author (books), a few authors (debates)
or a large number of authors (USENET), in writing and speech (e.g., books~vs.~debates), and in different languages 
(e.g., novels), indicating that the stretched exponential scaling is robust with regard
to sample size, number of authors, language mode, and language. \\

\noindent{\bf Conclusions.} The quest for statistical laws in language has been driven both
 by applications in text mining and document retrieval, and by the desire for
 foundational understanding of humans as agents and participants in the world. Taking
 texts as examples of extended discourse, 
we combined these research agendas by showing that word meanings are directly related to
their recurrence distributions via the permutability of concepts across discourse
contexts. Our model for generating long-term recurrence patterns of words, a bag-of-words
model 
with memory, is stationary and uniformly applicable to words of all parts of speech and
semantic types. A word's position along the range in the memory parameter in the model,
$\beta$, effectively captures its position in between a power-law and an exponential
distribution, thus capturing its degree of contextual anchoring.  Our results agree with
Ref.~\cite{hauser} in emphasizing 
both the specific ability to learn abstract operators
and the broader conceptual-intentional system as components in the human capability for
language and in its use in the flow of discourse.

Analogies between communicative dynamics and social dynamics more
generally are suggested by the recent documentation of  heavy-tailed distributions  in
many other human driven activities~\cite{barabasi,malmgren,vazquez}. They indicate that tracing linguistic
activities in the ever larger digital 
databases of human communications can be a most promising tool for tracing human and
social dynamics~\cite{watts}. The stretched exponential form for recurrence distributions that
derives from our model and the empirical finding it embodies are thus expected to also find 
applicability in other areas of human endeavor. \\

\section{Acknowledgments} 

We thank Pedro Cano and Seth Myers for piloting 
supplementary studies of word recurrences, Roger Guimer\`a and Seth Myers for sharing
web mining programs, Bruce Spencer for advice on the statistical analysis, Stefan Kaufmann for his 
assistance in developing the semantic coding scheme,  and 
Hannah Rohde for validating it. E. G. A. thanks the audiences at the Complex Systems \&
Language Workshop (The University of Arizona, April 2008) and at the NICO Colloquium (Northwestern
University, October 2008) for providing valuable feedback. This work was supported by JSMF
Grant No. 21002061 (J. B. P.) and NSF DMS-0709212 (A. E. M.).  

\section{Supporting Information}

The supporting information files are available at the PLoS ONE journal (open access) at
http://dx.plos.org/10.1371/journal.pone.0007678. 

{\bf Text S1:} Supplementary information on language analysis, 
statistical analysis, and counting models. Found at: doi:10.1371/journal.pone.0007678.s001

{\bf Table S1:} Detailed information on the statistical analysis of all 
words that were studied (six databases).  Found at: doi:10.1371/journal.pone.0007678.s002

\end{document}